\documentclass[10pt,twocolumn,letterpaper]{article}

\usepackage[pagenumbers]{cvpr} % To force page numbers, e.g. for an arXiv version
\usepackage{multirow}
\usepackage{graphicx}
\usepackage{empheq}
\usepackage{pifont}
\newcommand{\cmark}{\ding{51}}
\newcommand{\xmark}{\ding{55}}
\usepackage{booktabs}

\definecolor{cvprblue}{rgb}{0.21,0.49,0.74}
\usepackage[pagebackref,breaklinks,colorlinks,allcolors=cvprblue]{hyperref}

\def\paperID{} % *** Enter the Paper ID here
\def\confName{CVPR}
\def\confYear{2026}

\title{Linking Modality Isolation in Heterogeneous Collaborative Perception}

\author{Changxing Liu\textsuperscript{1}\thanks{Equal contribution} \quad Zichen Chao\textsuperscript{2}\footnotemark[1] \quad Siheng Chen\textsuperscript{1}\thanks{Corresponding author}
\\
\textsuperscript{1}Shanghai Jiao Tong University  \quad
\textsuperscript{2}Nanjing University of Science and Technology \quad 
\\
{\tt\small cx-liu@sjtu.edu.cn \quad zichen.chao@njust.edu.cn \quad sihengc@sjtu.edu.cn}
}

\begin{document}
\maketitle
\begin{abstract}
Collaborative perception leverages data exchange among multiple agents to enhance overall perception capabilities.
However, heterogeneity across agents introduces domain gaps that hinder collaboration, and this is further exacerbated by an underexplored issue: \textbf{modality isolation}. It arises when multiple agents with different modalities never co-occur in any training data frame, enlarging cross-modal domain gaps. Existing alignment methods rely on supervision from spatially overlapping observations, thus fail to handle modality isolation.
To address this challenge, we propose CodeAlign, the first efficient, co-occurrence-free alignment framework that smoothly aligns modalities via cross-modal feature-code-feature(FCF) translation. The key idea is to explicitly identify the representation consistency through codebook, and directly learn mappings between modality-specific feature spaces, thereby eliminating the need for spatial correspondence. Codebooks regularize feature spaces into code spaces, providing compact yet expressive representations. With a prepared code space for each modality, CodeAlign learns FCF translations that map features to the corresponding codes of other modalities, which are then decoded back into features in the target code space, enabling effective alignment. Experiments show that, when integrating three modalities, CodeAlign requires only 8\% of the training parameters of prior alignment methods, reduces communication load by 1024x, and achieves state-of-the-art perception performance on both OPV2V and DAIR-V2X dataset. Code will be released on https://github.com/cxliu0314/CodeAlign.
\end{abstract}
    
\section{Introduction}
\label{sec: 1}

Collaborative perception plays a pivotal role in intelligent systems such as connected autonomous vehicles and multi-robot collaboration. It enables agents to build a more comprehensive understanding of the environment by sharing perceptual information. 
% Current research in collaborative perception primarily operates under a homogeneous setting, focusing on issues such as pose alignment and communication compression to achieve more accurate and efficient collaboration (cite Where2Comm/...).
In real-world applications, however, vehicles from different manufacturers often exhibit heterogeneity, leading to significant domain gaps during feature-level collaboration. Heterogeneity includes different sensor types, sensor parameters, and perception models.
Late fusion bypasses heterogeneity by integrating detection outputs, but suffers from suboptimal performance, localization noise ~\cite{lu2022robust}, and communication latency ~\cite{wang2020v2vnet}. Traditional approaches~\cite{v2xvit, HM-ViT, hu2024communication} use shared fusion networks for collective training, learning the consistency of heterogeneous features corresponding to spatial positions. Further developments~\cite{luo2024plug} promote alignment supervised on feature contrastive loss under same scene. To enable extensible heterogeneous collaboration, some approaches~\cite{gao2025stamp, anonymous2025gtspace} generate standardized features to represent the environment, using contrastive learning to align modalities toward this common feature. These methods all rely on spatial-correspondence alignment. They require co-occurring training data, that is, data from different modalities must have shared observation within the same scene.
% While some studies (e.g., V2XViT, Calibrator, MPDA, PnPDA, PolyInter) have addressed heterogeneity within LiDAR-based systems, they fail to tackle the more fundamental challenge of cross-modal heterogeneity between LiDAR and camera. Latest research have utilized collective training (HMViT) or modalities extension (HEAL) to tackle the cross-modal heterogeneity problem.

\begin{figure}[t]
  \centering
  \includegraphics[width=0.47\textwidth]{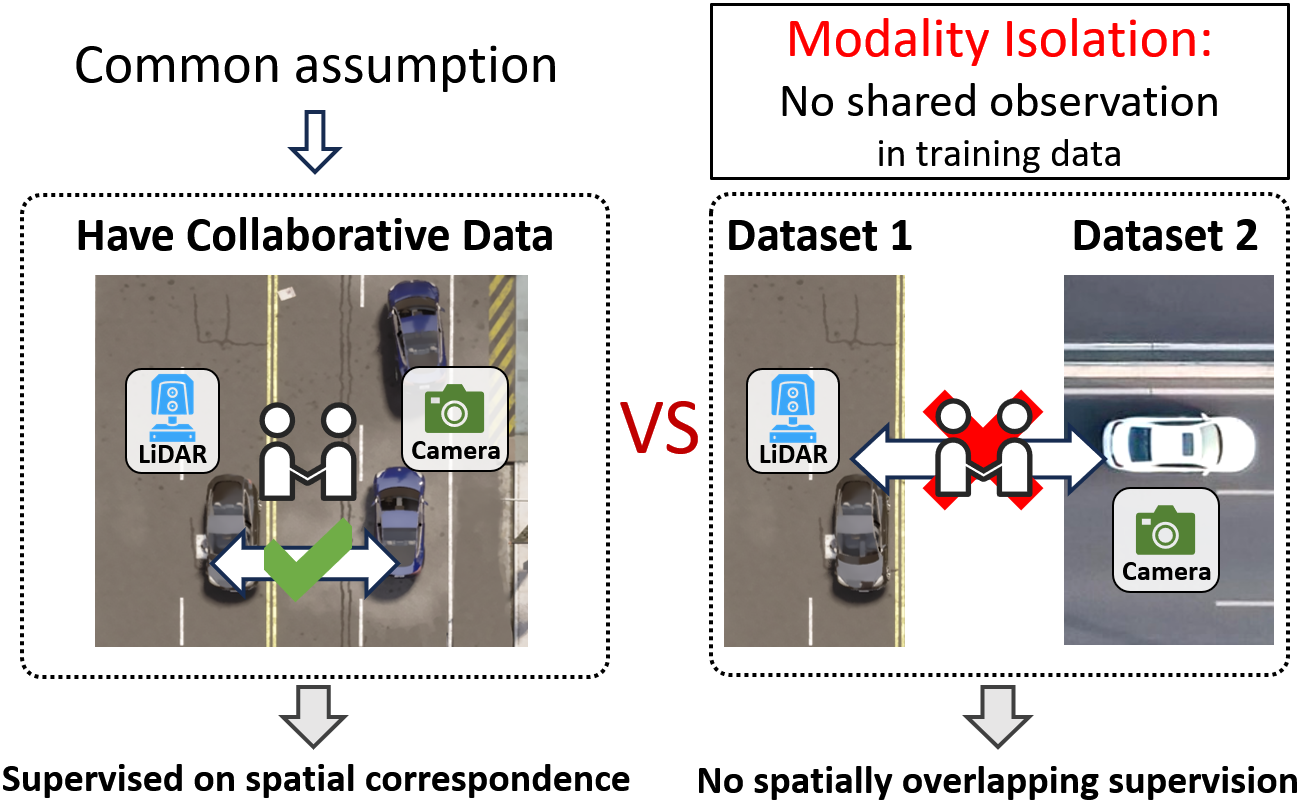}
  \vspace{-0.3cm}
  \caption{Illusion of modality isolation in heterogeneous collaborative perception.}
  \vspace{-0.7cm}
  \label{fig:intro}
\end{figure}

A critical yet understudied issue in multi-agent heterogeneous collaborative perception is modality isolation, as illustrated in Figure~\ref{fig:intro}. In real-world deployments, perception data of agents are typically collected by different institutions across diverse locations and time periods. Consequently, each dataset often covers only a limited subset of modalities. As a result, many modality pairs never co-occur in any recorded scene, meaning they lack shared observations, have never collaborated in the training data, and thus provide no shared spatially correspondence supervision. We refer to this situation as modality isolation.

Modality isolation significantly increasing the difficulty of achieving robust and generalizable alignment. When aligning two modality-isolated agents, the absence of shared observation in any frame makes it impossible to establish mutual supervision through correspondence of BEV features~\cite{gao2025stamp, HM-ViT}, nor can shared ground truth labels be utilized as reference to facilitate alignment ~\cite{anonymous2025gtspace}. For collectively trained fusion networks, it is feasible to alternately input single-modality data from isolated modalities; however, this significantly impairs perception performance for using modality-isolated features to train a shared backend. Although ~\cite{lu2024extensible} adapts to modality-isolated scenarios through a extension strategy trained on local data of each modalities, it is limited by the high training cost and inconvenience of retraining encoders.

Modality isolation post the following challenges: i) Heterogeneity arises from diverse factors, resulting in a wide spectrum of heterogeneous types. ii) Continuous technological advancements lead to the emergence of new modalities, which inevitably suffer from modality isolation with existing ones. iii) Datasets collected by different institutions are often subject to data privacy requirements, further restricting data accessibility. Under such conditions, the system must be capable of aligning numerous modalities efficiently, while maintaining extensibility and protecting data privacy.

To address these challenges, our core idea is to explicitly identify the representation consistency with codebook and directly learn mappings between modality-specific feature spaces. The feature spaces consist of spatial-irrelative, pixel-level representations, and the mappings between these spaces can be learned through representation consistency, thereby eliminating the reliance on spatial correspondence. To achieve reliable alignment, it is required that the feature space should be representative and condensed. Intuitively, codebook can construct compact yet expressive feature space, facilitating explicit and robust transformation.

Following this idea, we propose CodeAlign, the first efficient, co-occurrence-free alignment framework for heterogeneous collaborative perception. CodeAlign smoothly aligns modalities via cross-modal feature–code–feature(FCF) translation under modality isolation.
The framework operates in two stages:
i) Code space construction. To extract representative intermediate features, a inserted codebook is learned for each modality between their encoder and fusion network. In addition to providing condensed code-feature pairs, the codebook enables efficient communication by transmitting codebook indices instead of raw features. To further improve training efficiency, we introduce a group code space construction strategy that establishes a shared code space for non-isolated modalities, enhancing alignment accuracy while reducing alignment training effort.
ii) FCF translation. To align modalities, CodeAlign learns cross-modal FCF translation: features of the ego modality are translated into the target modality’s corresponding codes, which are then decoded back into features in the target code space, achieving indirect yet effective alignment. To ensure efficiency in both training and inference, we design a lightweight one-to-many Code Translator that supports translation to multiple code spaces.
Compared with existing methods, CodeAlign overcomes the challenge of modality isolation by learning representation consistency instead of spatial correspondence, while simultaneously reducing training costs and communication overhead.

We evaluate CodeAlign on the OPV2V~\cite{xu2022opv2v} and DAIR-V2X~\cite{yu2022dair} datasets and demonstrate significant improvements in efficiency. When integrating three modalities, CodeAlign uses only 8\% of the training parameters required by HEAL~\cite{Heal} and reduces communication load by a factor of 1024. Despite its discretization nature, CodeAlign also boosts perception performance on AP30 by up to 4.61\% on OPV2V and 12.08\% on DAIR-V2X compared to state-of-the-art alignment methods, highlighting its strong generalization ability.
Our contributions are as follows:

\begin{itemize}
\item We propose CodeAlign, the first efficient, co-occurrence-free alignment framework to resolve the modality isolation challenge in heterogeneous collaborative perception.
% , which replaces spatial correspondence with representation consistency as the supervision for cross-modal alignment.

\item CodeAlign introduces FCF translation to linking isolated modalities, enabling effective alignment while lowering training cost and reducing communication overhead.

\item Extensive experiments on OPV2X and DAIR-V2X datasets demonstrate the effectiveness of CodeAlign in improving perception performance while enhancing both training and communication efficiency.
\end{itemize}

\section{Related Works}
% \vspace{-0.3cm}

\subsection{Collaborative Perception}
% \vspace{-0.1cm}
Collaborative perception improves detection accuracy by leveraging shared sensory information across multiple agents and is commonly classified into early, intermediate, and late fusion strategies. Early fusion transmits raw sensor data, incurring high communication cost, while late fusion shares only bounding boxes, limiting performance and robustness due to feature loss ~\cite{wang2020v2vnet, lu2022robust}. Intermediate fusion ~\cite{tang2025cost, xu2025cosdh, GenerativeMapPriors} has gained popularity for achieving a favorable balance between performance and communication efficiency.
To advance research in multi-agent collaborative perception, OPV2V ~\cite{xu2022opv2v} provides simulated vehicle-to-vehicle collaboration, DAIR-V2X ~\cite{yu2022dair} offers real-world vehicle-infrastructure data, and RCooper ~\cite{hao2024rcooper} introduces adverse weather conditions for robustness evaluation. 
To mitigate communication bottlenecks, prior works~\cite{hu2022where2comm, hu2024pragmatic} reduce redundancy in transmitted features. Other approaches ~\cite{wei2023asynchrony, lei2022latency, zhang2025co} address communication disruptions or latency by exploiting historical interaction data or temporal context.

Despite these advances, most existing methods assume homogeneous sensor modalities and identical models across agents. In this work, we study heterogeneous collaborative perception, where agents may possess different sensor and model configurations. We further identify and investigate the underexplored challenge of modality isolation, arising when heterogeneous agents struggle to effectively align and fuse features due to no co-occurring data.

% \vspace{-0.3cm}
\subsection{Heterogeneous Collaborative Perception}
% \vspace{-0.1cm}
Heterogeneity in collaborative perception arises from differences in sensor modalities, sensor configurations, and perception model architectures. Early works focus on LiDAR-based heterogeneity: V2XViT~\cite{v2xvit} addresses spatial misalignment between vehicle and infrastructure; MPDA~\cite{xu2022bridging} and Calibrator~\cite{xu2022model} study heterogeneous LiDAR models; PnPDA~\cite{luo2024plug} further considers varying voxel sizes; and PolyInter~\cite{xia2025oneispenty} explores scalability. However, the domain gap between LiDAR and camera data constitutes a more significant challenge. HMViT~\cite{HM-ViT} proposes collective training of cross-modal models to bridge this gap. CodeFilling~\cite{hu2024communication} leverages a shared codebook in collective training, but is inconvenient for modality extension. STAMP~\cite{gao2025stamp} trains a protocol network from a certain modality to provide contrastive learning references, but requires that modality to be present in every data. GT-Space~\cite{anonymous2025gtspace} uses ground-truth-derived features as alignment anchors, yet assumes modalities share a common field of view.

However, the challenge of modality isolation arises from the fact that different modalities are rarely co-collected in the same scenes, leading to a lack of shared observation in training data. This limitation hinders spatial-correspondence alignment infeasible. HEAL~\cite{Heal} adapts to this situation by retraining encoders on local data, but suffers from high computational cost and inconvenience of an extra encoder. In this work, we propose CodeAlign, an efficient, co-occurrence-free alignment framework under modality isolation that leverages FCF translation to systematically address this challenge.
\section{Methodology}
\label{sec: 3}

\begin{figure*}[t]
  \centering
  \includegraphics[width=0.9\textwidth]{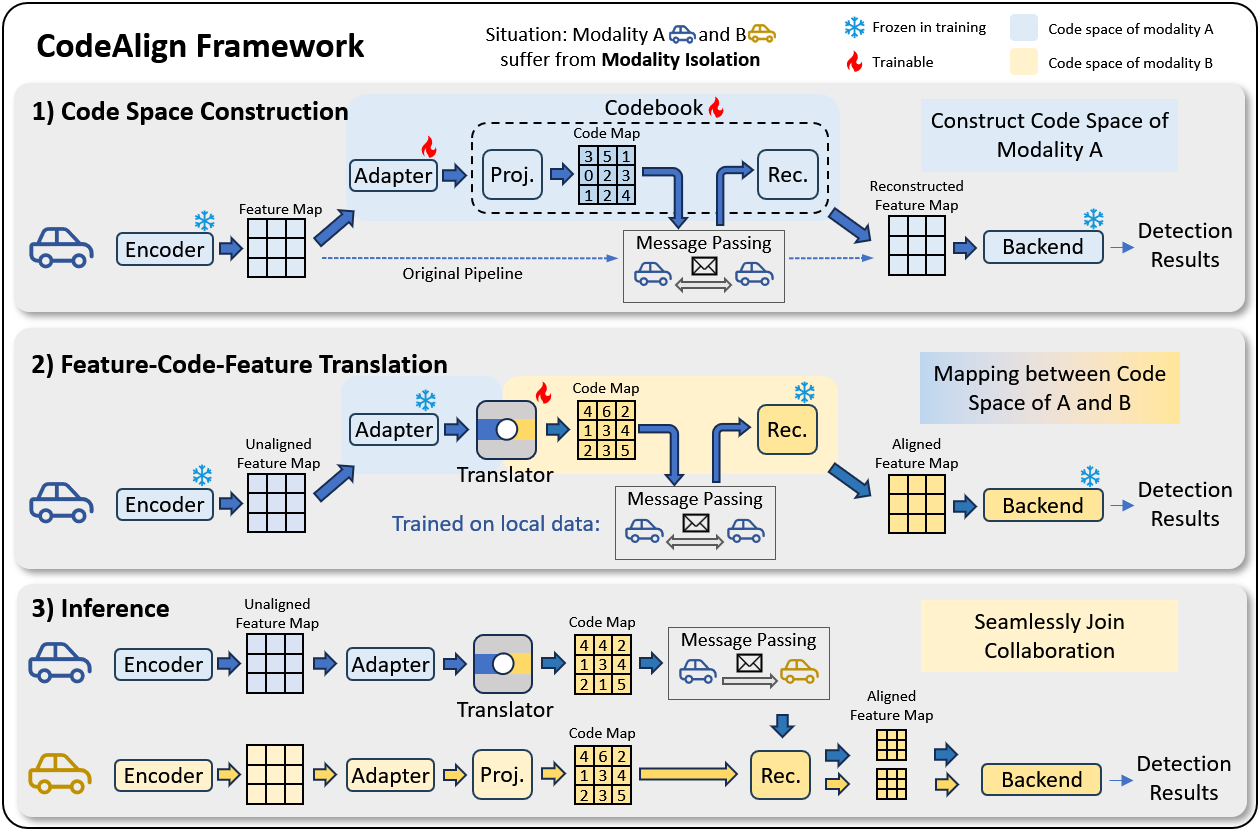}
  \vspace{-0.3cm}
  \caption{Overview of the CodeAlign framework. The two training stages are: (1) code space construction for each modality and (2) feature–code–feature (FCF) translation between modalities, both trained using single-modality data. Step (3) illustrates the inference process, in which modality A has never encountered modality B during offline training, yet can still collaborate with it seamlessly.}
  \vspace{-0.5cm}
  \label{fig:pipeline}
\end{figure*}

\subsection{Problem Formulation}
Consider a heterogeneous collaborative perception system with a set of agents $\mathcal{A}$, where each agent $i \in \mathcal{A}$ is equipped with modality $m_i$ and produces an observation $x_i^{m_i}$ for scene $s$ it participates in. Training data are collected across diverse scenes with diverse location and timestamp. Each recorded scene involves a subset $\mathcal{G}_s \subseteq \mathcal{A}$ of agents that simultaneously observe the environment. The total training dataset can be denoted as a collection of scene-observation pairs: $\bigcup_{s} \left\{ \left(s, \{x_i^{m_i}\}_{i \in \mathcal{G}_s} \right) \right\}$.

During conventional training, the system processes the co-occurring observations $\{x_i^{m_i}\}_{i \in \mathcal{G}_s}$ through a three-stage pipeline to produce detection outputs $\{\hat{y}_i\}_{i \in \mathcal{G}_s}$:
\begin{align}
    & F_i^{m_i} = E_{m_i}(x_i^{m_i}), i \in \mathcal{G}_s 
    \label{eq: 1}\\
    & F_{j \to i} = \Gamma_{j \to i}(\mathcal{M}_{j \to i}(F_j^{m_j})), j \in \mathcal{G}_s 
    \label{eq: 2}\\
    & \hat{y}_i = B_{m_i}\big( \{F_{j \to i}\}_{j \in \mathcal{G}_s} \big)
    \label{eq: 3}
\end{align}
where a modality-specific encoder $E_{m_i}$ extracts the intermediate feature $F_i^{m_i}$, and after message passing $\mathcal{M}_{j \to i}$, the spatial transformation $\Gamma_{j \to i}$ aligns agent $j$'s feature into agent $i$'s coordinate frame, finally the backend module $B_{m_i}$ fuses all transformed messages $\{F_{j \to i}\}_{j \in \mathcal{G}_s}$ to generate detection result $\hat{y}_i$.

Existing methods either use common ground-truth labels $y_s^{\text{gt}}$ or spatially overlapped intermediate features $\{F_i^{m_i}, F_{j \to i}^{m_j}\}$ to supervise alignment (listed in Section.~\ref{sec: compare method} in appendix). However, when two modalities never co-appear in any training scene, such supervision is absent, and existing approaches fail to establish meaningful alignment.

We refer to this situation as \textbf{modality isolation}. For any agent $i$, let $\mathcal{S}^{m_i} = \{ s \mid i \in \mathcal{G}_s \}$ denote the set of scenes covered by its modality $m_i$. Agents $i$ and $j$ experience modality isolation if
\begin{equation}
    \mathcal{S}^{m_i} \cap \mathcal{S}^{m_j} = \varnothing.
\end{equation}
Intuitively, because the two agents never appear in the same scene, none of their observations overlap in space, thus provide no supervision on spatial correspondence.

Our goal is to enable effective and efficient alignment under modality isolation without spatial-correspondence supervision. To this end, CodeAlign learns modality-specific code spaces from local data and constructs cross-modal FCF translation without requiring overlapping scene.

\subsection{CodeAlign Framework}
To address modality isolation, CodeAlign leverages the representative consistency wihtin each modality as the basis for alignment with the help of codebooks. By explicitly constructing feature spaces through codebooks and exposing their discrete representations, cross-modal alignment can be directly supervised via representation consistency, removing the reliance on co-occurring observational data.

To explicitly build feature spaces and ease cross-modal mapping, CodeAlign employs learnable codebooks to extract feature spaces as code spaces for each modality. This yields compact, semantic representations and reduces communication overhead in collaborative perception, as agents exchange code indices instead of dense features.

To integrate this design into the collaborative perception pipeline, we replace the conventional message passing step in Eq.~\ref{eq: 2} with a cross-modal feature-code-feature(FCF) translation mechanism. Specifically, given an intermediate feature $F_i^{m_i}$ from agent $i$, the translation performs:
% \vspace*{-0.3cm}
\begin{align}
    \mathbf{c}_i^{m_i} &= P_{m_i}\!\big( F_i^{m_i} \big), \quad \mathcal{B}^{m_i}\big[ \mathbf{c}_i^{m_i} \big] \in \mathcal{Z}^{m_i}, \\
    \mathbf{c}_i^{m_j} &= T_{m_i \to m_j}\!\big( F_i^{m_i} \big), \quad \mathcal{B}^{m_j}\big[ \mathbf{c}_i^{m_j} \big] \in \mathcal{Z}^{m_j}, \\
    \tilde{F}_j^{m_i} &= R_{m_i}\!\big( \mathcal{M}_{j \to i}(\mathbf{c}_j^{m_i}) \big), \quad \tilde{F}_j^{m_i} \in \mathcal{B}^{m_j}\big[ \mathbf{c}_i^{m_j} \big], \\
    F_{j \to i}^{m_i} &= \Gamma_{j \to i}\!\big( \tilde{F}_j^{m_i} \big).
\end{align}
% \vspace*{-0.3cm}
where $P_{m_i}$ denotes a projector that quantizes the intermediate feature $F_i^{m_i}$ into a discrete index map $\mathbf{c}_i^{m_i}$, whose corresponding embeddings $\mathcal{B}^{m_i}[\mathbf{c}_i^{m_i}]$ resides in the feature space $\mathcal{Z}^{m_i}$ defined by the codebook $\mathcal{B}^{m_i}$. $T_{m_i \to m_j}$ is a cross-modal translator that predicts an code map $\mathbf{c}_i^{m_j}$ for target modality $m_j$ from ego feature, achieving cross-modal feature-code translation. The reconstructor $R_{m_i}$ then decodes the transmitted codes into a dense feature $\tilde{F}_j^{m_i}$, which is mapped into $\mathcal{Z}^{m_j}$ with $\mathcal{B}^{m_i}$ as intermediary. The code-feature translation is achieved. The feature is transformed by $\Gamma_{j \to i}$ to yield the warped feature $F_{j \to i}^{m_i}$ for fusion.

The core insight of FCF translation lies in the direct correspondence between discrete indices and modality-specific embeddings: each index $\mathbf{c}_i^{m_j}$ uniquely specifies an embedding $\mathcal{B}^{m_j}[\mathbf{c}_i^{m_j}]$ within the target modality's native feature space $\mathcal{Z}^{m_j}$. The translator $T_{m_i \to m_j}$ predicts an index map directly, enabling the reconstructor $R_{m_j}$ to effectively decode a feature map composed of features just from the target modality's feature space. Consequently, the decoded features inherently align with the target modality's features, enabling seamless cross-modal correspondence.

The overall pipeline of CodeAlign is illustrated in Figure~\ref{fig:pipeline}. Step (1) performs \textit{code space construction}, where each modality learns its own codebook to represent perceptual features without modifying the original backbone. Within this stage, we introduce a group code space construction strategy that constructs a shared codebook for non-isolated modalities using their co-occurring data, thereby improving alignment quality while minimizing the number of required pairwise transformations. Step (2) carries out \textit{FCF translation}: cross-modal translators map ego features into the discrete codes of a target modality and decode them back into target features. To further improve efficiency and extensibility, a lightweight one-to-many Translator is designed to enable translation to multiple modalities. Step (3) shows the inference phase, in which a modality that has never co-occurred offline with another can still collaborate seamlessly through the learned code spaces and translators.

\subsubsection{Code Space construction}

The purpose of code space construction is to explicitly construct representative feature space of the modality-specific intermediate encoded features, and prepare the feature space for later cross-modality alignment. 
In CodeAlign, a learnable codebook is employed to extract representative feature space as code space of a modality, which yields compact and communication-efficient representations. The ego vehicle’s collaborative perception pipeline operates as follows: raw sensor data is first encoded into bird’s-eye-view (BEV) features; these features are then quantized into the code space by assigning each spatial location to its nearest codebook embedding, resulting in a compact code map composed of discrete indices. During communication, only this code map is transmitted, significantly reducing bandwidth requirements. Upon receiving code maps from neighboring agents, the ego vehicle decodes them back into reconstructed BEV features, which are subsequently fused by the backend module to produce final detection outputs.

Based on the codebook $C$, agent can replace the encoded feature map $F_{k}$ with a series of code indices $I_{k}$, forming a compact code map. For each BEV location $(h, w)$, the code index is computed by calculator $\Phi$ as,

\vspace*{-0.5cm}
\begin{equation}
    \Phi: I_{[h, w]} = \arg \min_{\ell} \left\| (\mathcal{P}(F))_{[h, w]} - C[\ell] \right\|_2^2
    \label{eq:6}
\end{equation}
\vspace*{-0.5cm}

The code map is used for message passing and is decoded by $\mathcal{R}$ to reconstruct the aligned feature map, as the aligned features maps are composed of deterministic discrete features in the code space. The transmitted intermediate feature is compressed from $H*W*C$ to $H*W*log2(D)$, where $D$ denotes the codebook size, significantly reducing communication bandwidth. 

During training, code space construction is performed in an incremental, plug-in manner. The original perception pipeline remains unchanged: we insert a lightweight adapter and a learnable codebook between the encoder and the backend, and we update only these inserted modules while freezing both the encoder and the backend. This design allows the code space to be built without interfering with the normal operation of the non-collaborative pipeline, and by preserving the original encoder, it avoids the loss of feature extraction accuracy that may arise in codebook-based end-to-end retraining. Furthermore, fixing both the encoder and the backend drastically reduces training parameters compared with end-to-end methods and HEAL~\cite{Heal}.
% In perception systems, the encoder extracts features from raw inputs, and is deeply coupled with downstream tasks. Arbitrary modifications to the encoder can degrade overall system performance and stability. Moreover, encoders for certain modalities (such as vision-based cameras) are often parameter-intensive.
% HEAL~\cite{Heal} achieves cross-modal alignment by retraining the encoder, which not only disrupts the consistency of the pre-trained perception pipeline but also incurs significant additional training overhead. In contrast, CodeAlign preserves the original encoder in a frozen state throughout training. Instead, it introduces a lightweight ResNet-based adapter and a learnable codebook only at an intermediate layer, thereby maintaining the integrity and modularity of the overall system. As shown in Fig.~\ref{fig:pipeline} (a), for single modality code space construction, we also keep the original backend fixed to minimize computational cost.

\textbf{Group Code Space Construction.} For non-isolated modalities, group code space construction enables them to jointly learn a shared code space, which is essential for improving training efficiency. Without grouping, each modality would need to be aligned pairwise with every other modality, resulting in quadratic growth in training cost. By contrast, grouping reduces this complexity, decreasing the total number of required cross-modal alignments. Moreover, it fully leverages collaborative data to strengthen alignment quality.

Since the encoded BEV features from different modalities are quantized into the same shared code space, effective alignment is naturally achieved. For example, features corresponding to a vehicle observed by different modalities may be mapped to the same codebook embedding, yielding identical decoded representations, which facilitates seamless fusion. For group code space construction, each modality is assigned its own adapter to facilitate alignment. The backend network is shared across modalities and remains trainable, as it must adapt to the new representation domain, and is well trained by collaborative data.

\textbf{Loss Function.} To accelerate alignment and ensure consistent feature representations across agents, we supervise three objectives: object detection, fusion learning, and inter-agent feature similarity. The overall loss is defined as:

\vspace*{-0.3cm}
\begin{equation}
\begin{split}
    L = {}& L_{\text{det}} \left( \widehat{\mathcal{O}}_i, \mathcal{O}_i^0 \right) + L_{\text{pyramid}} \\
    &+ \lambda \sum_{\substack{k, j \in \mathcal{G}_s ,\ m_k \neq m_j}} 
       L_{\text{sim}}\big(F_{k \rightarrow i}, F_{j \rightarrow i}\big),
\end{split}
\end{equation}
\vspace*{-0.3cm}

where $L_{\text{det}}(\cdot)$ denotes the detection loss, $\mathcal{O}_i^0$ and $\widehat{\mathcal{O}}_i$ represent the ground-truth and predicted object states for agent $i$, respectively. $L_{\text{pyramid}}$ is the pyramid loss from HEAL~\cite{Heal}, as we employ pyramid fusion as fusion net. $L_{\text{sim}}(\cdot)$ is similarity loss, which enforces feature consistency among collaborating agents. The similarity loss is applied over all pairs of agents $k, j \in \mathcal{G}_s$ that use different modalities ($m_k \neq m_j$), encouraging their features to be consistent from the perspective of the ego receiver $i$. We adopt the Smooth L1 loss between pairwise aligned features, which is less sensitive to outliers than the L2 loss. During the optimization, the network parameters and the codebook are updated simultaneously. Note that for single modality code space construction, no similarity loss is calculated.

\begin{table*}[]
\centering
\caption{Modality settings in the experiment. The network after LSS is its backbone.}
\vspace{-0.3cm}
\label{tab:modalities}
\resizebox{\textwidth}{!}{%
\begin{tabular}{l|ccccccc}
\hline
Modality & \textbf{m1} & \textbf{m2} & \textbf{m3} & \textbf{m4} & \textbf{m5} & \textbf{m6} & \textbf{m7} \\ \hline
Sensor Type & LiDAR(64-beam) & LiDAR(64-beam) & LiDAR(64-beam) & LiDAR(32-beam) & LiDAR(64-beam) & Camera & Camera \\
Encoder & \begin{tabular}[c]{@{}c@{}}Point-Pillar\\ (voxel: 0.4m)\end{tabular} & \begin{tabular}[c]{@{}c@{}}Second\\ (voxel: 0.1m)\end{tabular} & \begin{tabular}[c]{@{}c@{}}VoxelNet\\ (voxel: 0.4m)\end{tabular} & \begin{tabular}[c]{@{}c@{}}Point-Pillar\\ (voxel: 0.4m)\end{tabular} & \begin{tabular}[c]{@{}c@{}}Point-Pillar\\ (voxel: 0.3m)\end{tabular} & \begin{tabular}[c]{@{}c@{}}LSS\\ (ResNet-101)\end{tabular} & \begin{tabular}[c]{@{}c@{}}LSS\\ (EfficientNet)\end{tabular} \\ \hline
\end{tabular}%
}
\vspace{-0.5cm}
\end{table*}

\begin{table}
    \centering
    \caption{When m1 and m6 are isolated in their modalities, Pyramid Fusion~\cite{Heal} exhibits degraded perception performance.}
    \vspace{-0.2cm}
    \resizebox{0.32\textwidth}{!}{%
    \begin{tabular}{l|lll}
        \hline
        Modality Type    & AP30  & AP50  & AP70  \\ \hline
        Non-Isolated & \textbf{89.96} & \textbf{88.52} & \textbf{80.88} \\
        Isolated     & 82.36 & 80.51 & 65.67 \\ \hline
    \end{tabular}
    }
    \label{tab:isolation}
\vspace{-0.5cm}
\end{table}

\subsubsection{Feature-code-feature(FCF) Translation}
\label{sec:3.2.2}

Since each modality has established its own code space to represent modality-specific features, it requires a mechanism to translate between these heterogeneous code spaces. Within the CodeAlign framework, translation can occur between dense features or code maps. Among the feasible strategies, dense-to-dense translation offers lossless transformation but incurs high computational cost and forfeits the bandwidth efficiency advantage. Conversely, code-to-code translation suffers from excessive quantization error due to its highly discrete nature, causing significant information loss and degraded alignment performance. The dense-to-code approach achieves the optimal balance: it preserves the low-bandwidth communication benefits while maintaining reconstruction fidelity within acceptable limits, making it the practical solution for FCF translation.

Given an encoded dense feature from a source group, the translator maps it into a code map defined by the target group’s codebook. This compact representation is then transmitted and decoded using the target group’s codebook decoder, allowing the feature to be reconstructed in the target group’s code space and seamlessly integrated into its collaborative perception pipeline. 

The direct implementation of a code translator is a simple one-to-one translator, where each pair of groups trains a translator for each other. However, in scenarios involving collaboration among multiple modalities, which are usual, the conventional one-to-one translation paradigm suffers from significant drawbacks: a complex training process and high inference memory usage. The training complexity arises because a dedicated translator must be trained for every possible pair of groups, and often in both directions. During inference, the system must load all trained translators into memory to be prepared for potential collaboration with any modality, leading to substantial memory overhead. To address this issue, we propose a lightweight one-to-many Code Translator, which equipped with modality-specific multi-heads. We train a shared backbone with multiple output heads, each dedicated to a specific target modality. The model backbone is implemented with stacked ConvNeXt blocks, and before and after the backbone, two channel converters are placed to transform different input channels. During training, only the translator is trainable, with encoder, adapter, codebook and backend frozen. In addition, we design a data balancing strategy to dynamically adjust the proportion of training data according to the loss changes of different targets, which promotes balanced spatial learning.
Notably, when translation is required between only two groups, the model can degenerate into a standard one-to-one translator without relying on multi-heads.

\textbf{Local Data Training.} To avoid the defects of modality isolation, we design a training protocol that relies exclusively on local data: the source modality processes its own data through its encoder and the code translator, with the generated code map fed directly into the target group's reconstructor and backend. The detection loss computed on the target backend's output serves as the supervision, explicitly encouraging the translator to produce features that align with the target group's code space representation. This approach requires no external data transmission, achieves effective alignment by local training with the ego modality's data, and fully complies with data privacy regulations while enabling cross-institutional collaboration.

% embedding training for new groups

\section{Experiments}

\subsection{Experimental Settings}

\begin{table*}[]
\centering
\caption{Perception performance on OPV2V. The first row explains the tested collaboration scenarios of different isolated modalities, from a total of 2 to 3 cars. TP: training parameters for alignment. At last column, we show the average one-shot communication load.}
\vspace{-0.2cm}
\label{tab:opv2v}
\resizebox{\textwidth}{!}{%
\begin{tabular}{l|cccc|cccc|cccc|c}
\hline
\multicolumn{1}{c|}{} & \multicolumn{4}{c|}{m1 + m2} & \multicolumn{4}{c|}{m1 + m7} & \multicolumn{4}{c|}{m1 + m7 + m2} &  \\
\multicolumn{1}{c|}{\multirow{-2}{*}{Method}} & AP30 & AP50 & AP70 & TP/M & AP30 & AP50 & AP70 & TP/M & AP30 & AP50 & AP70 & TP/M & \multirow{-2}{*}{Comm Load} \\ \hline
No Collaboration & 81.18 & 79.44 & 68.26 & 0 & 81.18 & 79.44 & 68.26 & 0 & 81.18 & 79.44 & 68.26 & 0 & 0 \\
No Alignment & 78.33 & 76.35 & 61.62 & 0 & 77.59 & 76.42 & 66.01 & 0 & 72.91 & 71.58 & 59.66 & 0 & 32MB \\
Late Fusion & 92.73 & 91.46 & 79.34 & 0 & 84.47 & 80.12 & 62.40 & 0 & 88.24 & 85.02 & 68.45 & 0 & 0.5KB \\ \hline
Pyramid Fusion~\cite{Heal} & 91.98 & 91.36 & 85.92 & 6.5 & 82.58 & 81.13 & 69.26 & 20.5 & 83.95 & 82.93 & 68.91 & 21.4 & 32MB \\
HMViT~\cite{HM-ViT} & 70.31 & 69.74 & 60.86 & 33.8 & 84.22 & 79.55 & 54.97 & 41.3 & 86.96 & 83.98 & 64.39 & 62.7 & 32MB \\
CodeFilling~\cite{hu2024communication} & 91.98 & 91.11 & 80.68 & 5.3 & 78.37 & 74.27 & 44.03 & 5.3 & {\color[HTML]{1F2329} 81.37} & {\color[HTML]{1F2329} 80.54} & {\color[HTML]{1F2329} 60.81} & 5.3 & \textbf{0.03MB} \\
HEAL~\cite{Heal} & 93.02 & 92.10 & \textbf{86.18} & 1.0 & 82.45 & 81.03 & 71.70 & 15.0 & 87.8 & 86.98 & \textbf{79.89} & 16.0 & 32MB \\
CodeAlign & \textbf{93.39} & \textbf{92.67} & 85.56 & \textbf{0.8} & \textbf{87.19} & \textbf{85.3} & \textbf{72.1} & \textbf{0.8} & \textbf{89.77} & \textbf{88.59} & 77.73 & \textbf{1.3} & \textbf{0.03MB} \\ \hline
\end{tabular}%
}
\vspace{-0.2cm}
\end{table*}

\begin{table*}[t]
\centering

% Left table: DAIR-V2X (60% of text width)
\begin{minipage}{0.66\textwidth}
\centering
\caption{Performance on DAIR-V2X dataset. We assign modality m1 to the ego vehicle and evaluate collaboration with different infrastructure modalities.}
\vspace{-0.2cm}
\label{tab:dair}
\resizebox{\textwidth}{!}{%
\begin{tabular}{l|ccc|ccc|ccc}
\hline
\multicolumn{1}{c|}{\multirow{2}{*}{Method}} & \multicolumn{3}{c|}{m1 + m2} & \multicolumn{3}{c|}{m1 + m6} & \multicolumn{3}{c}{m1 + m7} \\
\multicolumn{1}{c|}{} & AP30 & AP50 & AP70 & AP30 & AP50 & AP70 & AP30 & AP50 & AP70 \\ \hline
No Collaboration & 24.68 & 17.20 & 4.70 & 24.68 & 17.20 & 4.70 & 24.68 & 17.20 & 4.70 \\
Late Fusion & 53.33 & 33.15 & 1.45 & 25.99 & 17.13 & 0.95 & 27.13 & 13.2 & 0.6 \\
HEAL~\cite{Heal} & 73.7 & 67.21 & 44.76 & 63.68 & 57.08 & 41.05 & 64.37 & 57.87 & 42.4 \\
CodeAlign & \textbf{82.03} & \textbf{77.37} & \textbf{57.84} & \textbf{72.12} & \textbf{64.86} & \textbf{45.45} & \textbf{74.09} & \textbf{65.25} & \textbf{45.84} \\ \hline
\end{tabular}%
\vspace{-0.7cm}
}
\end{minipage}
\hfill
% Right table: Ablation (40% of text width)
\begin{minipage}{0.31\textwidth}
\centering
\caption{Ablation in group code space construction with m1+m6. CB: codebook(16); Fix: frozen encoder; Ada.: with adapter.}
\vspace{-0.2cm}
\label{tab:ablation group}
\resizebox{\textwidth}{!}{%
\begin{tabular}{cccc|ccc}
\toprule
CB & Fix & Ada. & $L_{sim}$ & AP30 & AP50 & AP70 \\
\midrule
   &      &      &           & 89.30 & 88.03 & 80.44 \\
\cmidrule(lr){1-7}
\cmark &      &      &           & \textbf{89.54} & \textbf{87.74} & 77.87 \\
\cmark & \cmark & \cmark &           & 88.37 & 86.83 & 78.09 \\
\cmark & \cmark & \cmark & \cmark    & 89.04 & 87.54 & \textbf{79.63} \\
\bottomrule
\end{tabular}%
}
\end{minipage}
\end{table*}

\begin{table}
    \centering
    \setlength{\tabcolsep}{5pt}
\vspace{-0.4cm}
    \caption{Performance of different align method for modalities that have collaborative data (consider m1 and m6). Results show the benefit of group code space construction.}
    \vspace{-0.1cm}
    \resizebox{0.4\textwidth}{!}{%
    \begin{tabular}{l|ccc}
        \hline
        \multicolumn{1}{c|}{Align Method}                & AP30  & AP50  & AP70  \\ \hline
        Group Code Space Construction       & \textbf{89.04} &\textbf{87.54} & \textbf{79.63} \\
        FCF Translation       & 87.96 & 86.00 & 72.92 \\ \hline
    \end{tabular}
    }
    \label{tab:group}
    \vspace{-0.3cm}
\end{table}

\begin{table}[]
    \centering
    \setlength{\tabcolsep}{3pt}
    \caption{Ablation on model structures of cross-modal alignment with 4 isolated modalities. * means using data balance strategy.}
    \label{tab:ablation inter}
    \vspace{-0.1cm}
    \resizebox{0.47\textwidth}{!}{%
    \begin{tabular}{l|ccccc}
    \hline
    \multicolumn{1}{c|}{Alignment Method} & AP30 & AP50 & AP70 & TP/M & TP(n modalities) \\ \hline
    Backward Alignment~\cite{Heal} & 82.28 & 81.28 & 74.66 & 17.7 & $\sum_n Enc$ \\
    One-to-one Code Translator & \textbf{90.30} & \textbf{88.61} & \textbf{79.01} & 1.9 & $(0.2M)n^2$ \\
    Multi-head Code Translator & 90.17 & 88.51 & 78.33 & 1.5 & $(0.5M)n$ \\
    Multi-head Code Translator* & 90.25 & 88.58 & 78.42 & 1.5 & $(0.5M)n$ \\ \hline
    \end{tabular}%
    }
    \vspace{-0.2cm}
\end{table}

\textbf{Dataset.} CodeAlign is evaluated on the simulated OPV2V~\cite{xu2022opv2v} dataset and the real world dataset DAIR-V2X~\cite{yu2022dair}. OPV2V is a large-scale multi-modal cooperative V2V perception dataset collected in CARLA~\cite{Dosovitskiy17} and OpenCDA~\cite{xu2021opencda}, where each scenario contains multiple connected AVs and each AV is equipped with 1 LiDAR and 4 monocular cameras. DAIR-V2X is a large-scale, real-world vehicle-infrastructure cooperative perception dataset featuring synchronized camera and LiDAR sensor data from both ego vehicles and roadside units across diverse urban and highway scenarios.

\noindent
\textbf{Modality Settings.} Following HEAL's design for modality variation, seven modalities are used in the experiments, as summarized in Tab.~\ref{tab:modalities}. The selected modalities include LiDAR and camera—two distinct sensor types with a significant domain gap—each equipped with dedicated encoders to ensure diversity in representation. We present results for a subset of modalities in the main text; additional results and more experiments are provided in the appendix Section.~\ref{sec: more exp}.

\noindent
\textbf{Implementation Details.} We employ the pyramid fusion network as the fusion net in backend. The adapter is implemented as a stack of four ResNet blocks with 3×3 convolutions. A codebook size of 16 is used across all experiments. Experiments are conducted on NVIDIA GeForce RTX 3090 GPUs. All models are trained using the Adam optimizer with an initial learning rate of 0.002. Code space construction is trained for up to 50 epochs, and FCF translation is trained for up to 30 epochs. The weight of the smooth L1 loss is set to 0.1. Training is conducted within the spatial range $x \in [-102.4m, +102.4m], y \in [-102.4m, +102.4m]$. 

\noindent
\textbf{Validation Settings.} For the main experiments on OPV2V and DAIR-V2X, we follow the settings of HEAL~\cite{Heal}, taking the first agent as the ego and incrementally adding new modalities; the agent order follows a fixed mapping to the car IDs in the validation dataset.
For other experiments involving two collaborating modalities, we alternately treat each modality as the ego vehicle and the other as the neighbors, and report the average result to more comprehensively reflect their alignment between each other. When reporting training parameters, we assume that the single-modality pipeline is pretrained and count only the parameters involved in alignment training. More details in Section.~\ref{sec: imple detail}.

\subsection{Quantitative Results}
\textbf{Impact of Modality Isolation.} Modality isolation largely degrades end-to-end training performance. Table~\ref{tab:isolation} shows the performance of pyramid fusion on dataset include m1 and m6. When the two modalities are not isolated and have shared observations, training yields high performance. In contrast, under modality isolation, m1 and m6 have not collaborated in training data, and the model is trained by alternately feeding data from each modality. The lack of shared supervision hinders effective feature alignment, resulting in a 15.21\% drop in AP70.

\begin{figure}
  \centering
  % \vspace{-0.4cm}
  \includegraphics[width=0.47\textwidth]{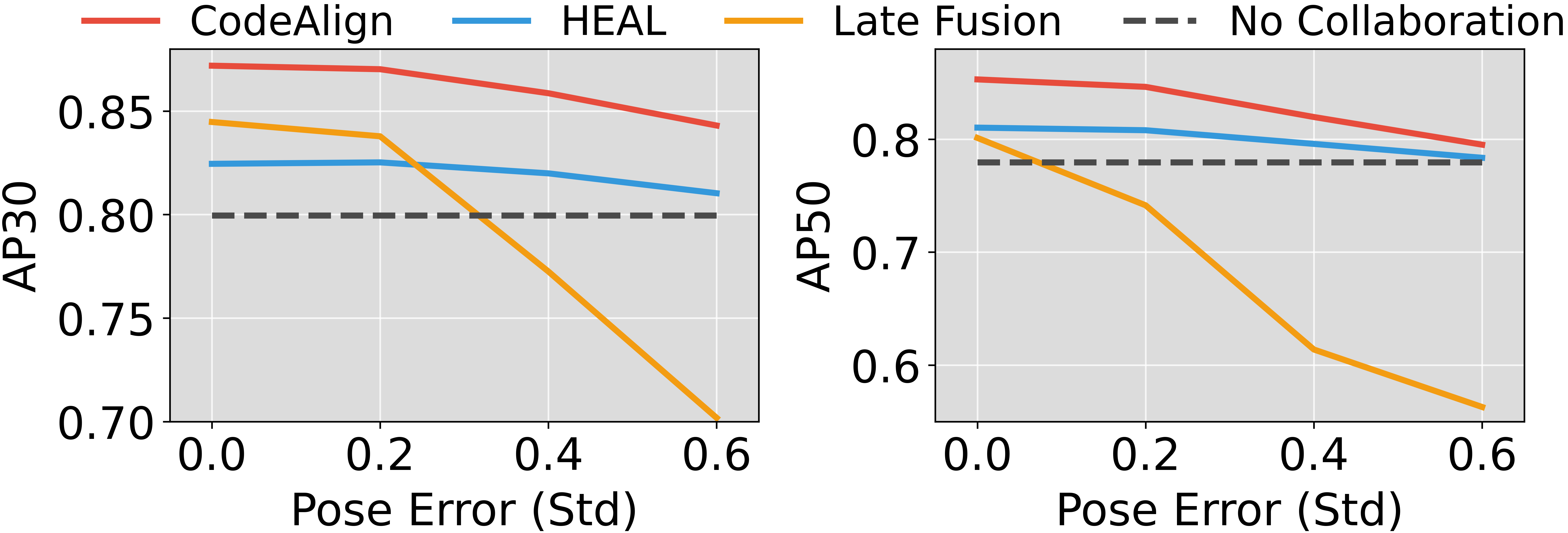}
  \vspace{-0.2cm}
  \caption{Robustness under pose error under scene m1+m7.}
  \vspace{-0.7cm}
  \label{fig:pose}
\end{figure}

\noindent
\textbf{Performance on OPV2V.} Table~\ref{tab:opv2v} illustrates the performance of CodeAlign compared to other baseline methods on the OPV2V dataset. We evaluate three modality-isolated collaboration scenarios, with varying modalities and agent numbers.
Three end-to-end fusion methods, Pyramid Fusion, HM-ViT and CodeFilling, were trained using alternative data for each modality. Pyramid Fusion only works for similar LiDAR modalities (m1+m2) through shared backend, but is ineffective across LiDAR–camera pairs. HM-ViT, lacking cross-modal co-occurring data for attention module training, yields negative gains. For CodeFilling, all modalities share a codebook of size 16, but the performance degrades due to the strong heterogeneity between LiDAR and camera. Late Fusion, which directly merges bounding boxes, adapts to modality isolation but falls short in accuracy, lagging behind CodeAlign by an average of 8.4\% in AP70.
Comparing CodeAlign with HEAL, CodeAlign's extraction of representative discrete features leads to an average improvement of 2.36\% and 2.15\% over HEAL in AP30 and AP50. Furthermore, CodeAlign significantly reduces training overhead with a 8\% training parameter of HEAL's in the three-vehicle scenario, leading to easier extension. Moreover, CodeAlign drastically cuts communication load to 1/1024th of intermediate fusion methods, enhancing the applicability of collaborative perception. Despite its discrete feature representation causing slightly lower AP70 scores than HEAL in some cases, CodeAlign maintains comparable overall performance. 
Other heterogeneous cooperative perception methods that do not support modality isolation are not included in this comparison and are discussed in Section~\ref{sec: compare method} of the appendix.

\noindent
\textbf{Robustness under pose error.} To evaluate the robustness of different methods in chaotic environments, we conduct experiments with pose error in the m1+m7 scenario. As shown in Figure~\ref{fig:pose}, the accuracy of late fusion drops rapidly as pose error increases, even falling below the no-collaboration baseline. In contrast, both CodeAlign and HEAL maintain effective collaboration, with CodeAlign consistently outperforming HEAL, demonstrating the strongest robustness to spatial noise.

\noindent
\textbf{Performance on DAIR-V2X.} On the more challenging real-world DAIR-V2X dataset shown in Table~\ref{tab:dair}, CodeAlign significantly outperforms other baselines when collaborating with different isolated modalities. Both Late Fusion and HEAL suffer from ambiguous and poor feature representations, whereas CodeAlign enhances semantic representation by explicitly extracting compact discrete features.

\noindent
\textbf{Benefit of Group Code Space construction.} Table~\ref{tab:group} shows the performance on non-isolated modality pairs m1 and m6, and illustrates the benefits of group code space construction. When modalities share a collaborative dataset, group code space construction leverages actual collaborative data for feature fusion and achieves 6.71\% higher AP70 than FCF translation approach.

\noindent
\textbf{Group Code Space Construction Strategy.} The first strategy in Table~\ref{tab:ablation group} is the end-to-end baseline without codebook, serving as an approximate upper bound. Adding a codebook preserves AP30/50 but drops AP70 by 2.57\% to 77.87\%, indicating quantization harms fine-grained accuracy. CodeAlign mitigates this by freezing the encoder and inserting an adapter, recovering AP70 to 78.09\%. With the supervision of the similarity loss, AP70 further improves to 79.63\% while maintaining high AP30/50. Our strategy closely approach lossless performance with significantly reduced communication cost.

\noindent
\textbf{Cross-modal Alignment Strategy.} In Table~\ref{tab:ablation inter}, we evaluate alignment from modality m2 to m1/m6/m7 for code translators, reporting average performance across scenes where m1/m6/m7 serve as the ego and m2 as the neighbor. Backward alignment underperforms in this setting and incurs high training cost proportional to the encoder size. One-to-one code translator achieve the best AP scores but require training parameters that grows quadratically with the number of modalities, limiting scalability. In contrast, the multi-head translator matches their performance and suffers only a 0.10\% drop in AP50, while scaling linearly in parameters, drastically reducing training complexity and is practical for large-scale deployment.

\subsection{Qualitative Results}
\begin{figure}[t]
  \centering
  \includegraphics[width=0.47\textwidth]{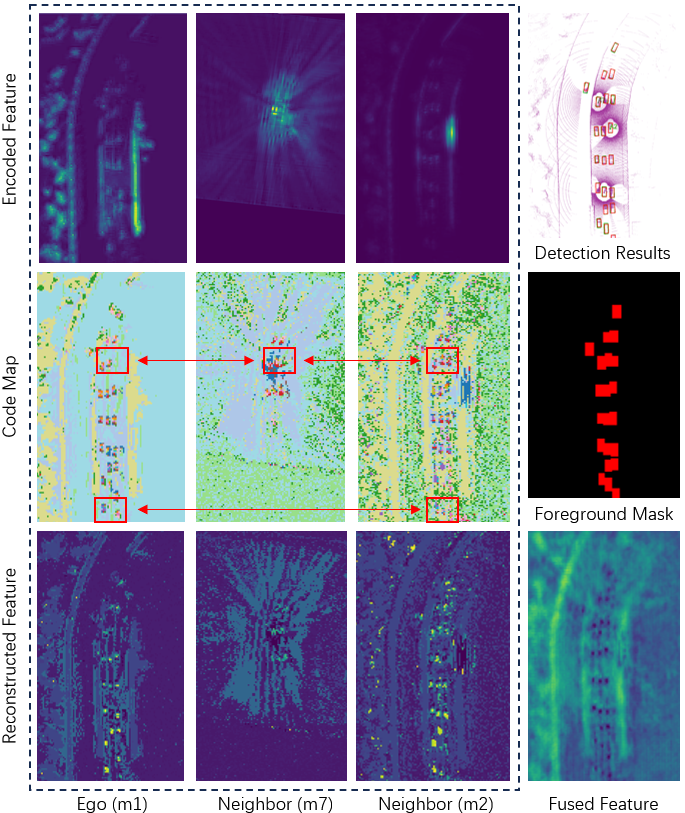}
  \vspace{-0.2cm}
  \caption{Visualization of BEV features under scene m1+m7+m2, illustrating the strong alignment ability of FCF translation.}
  \vspace{-0.7cm}
  \label{fig:vis}
\end{figure}

Figure~\ref{fig:vis} presents visualization results for the m1+m7+m2 scenario, illustrating the effectiveness of FCF translation in cross-modal alignment. Although feature of the camera-based modality m7 has inherently weaker representational capacity, after feature–code translation and mapping into the code space of m1, its spatial cues are explicitly captured. Through subsequent code–feature translation, the code map is translated to feature map that is well aligned with the feature space of m1, thereby smoothly achieves collaboration with m1. Similarly, the initially encoded feature of m2 exhibit a clear domain gap with respect to m1, but via FCF translation, the final feature become well aligned with feature of m1, enabling more advanced collaborative perception. The red boxes highlight objects that are missed by the ego vehicle but successfully recovered through collaboration with surrounding agents, demonstrating improved perception performance.

\section{Conclusion}
In this work, we address modality isolation — a critical yet underexplored challenge in heterogeneous collaborative perception, where the lack of co-occurring data across certain modalities exacerbates domain gaps and undermines conventional alignment strategies based on spatial correspondence. To tackle this, we propose CodeAlign, the first efficient, co-occurrence-free alignment framework that rely on establish explicit representation consistency to avoid lack of shared observation. By leveraging feature-code-feature translation, CodeAlign enables scalable, and communication-friendly collaboration across diverse agents with isolated modalities. 

\noindent
\textbf{Limitations.}  Our evaluation is limited by the modality diversity in datasets, preventing large-scale testing of group-wise alignment.

{
    \small
    \bibliographystyle{ieeenat_fullname}
    \bibliography{main}
}

% WARNING: do not forget to delete the supplementary pages from your submission 
\clearpage
\setcounter{page}{1}
\maketitlesupplementary

\section{Implementation Details}
\label{sec: imple detail}
\subsection{Validation}
We adopt two evaluation types in our experiments, detailed as follows:

Type 1 (Tables~\ref{tab:opv2v}, ~\ref{tab:dair}, ~\ref{tab:group_inter_1}, ~\ref{tab:group_inter_2}, following HEAL):  
For each scene in the validation set, vehicles are first assigned sequential IDs (1-4). The modalities to be evaluated are then mapped onto these vehicles in order, with the first modality always assigned to the ego vehicle (ID = 1). For example, in a scenario labeled 'm1 + m2', modality \(m_1\) is assigned to vehicle 1 (ego), and \(m_2\) to vehicle 2 (neighbor); higher-numbered vehicles remain unmapped, so there are up to 2 cars in this scenario. Note that this fixed ordering affects data distribution, that modalities appearing later in the sequence occur less frequently across scenes. In scenes with fewer than four vehicles, unmapped IDs are simply omitted.

Type 2 (all other tables):  
To more comprehensively evaluate performance in multi-agent collaborative settings, we adopt: for a two-modality collaboration scenario mi + mj, we assign mi to the ego vehicle (ID = 1) and set all other collaboratable vehicles to modality mj. For multi-modality scenarios (as in Table~\ref{tab:ablation inter}), we fix the ego modality (m2) and iteratively assign each remaining modality as mj to all non-ego vehicles, performing inference multiple times and averaging the results to assess pairwise cross-modal alignment. For single-modality evaluation, all collaboratable vehicles are assigned the same modality and participate jointly in the collaboration.

For OPV2V, both training and evaluation are conducted within the spatial range: $x \in [-102.4m, +102.4m], y \in [-102.4m, +102.4m]$. And for DAIR-V2X, both ranges are: $x \in [-102.4m, +102.4m], y \in [-51.2m, +51.2m]$

\subsection{Module Structure and Parameter Statistics}

\begin{table}[h]
\centering
\caption{Parameter for modality-specific encoders and modality-agnostic modules.}
\label{tab:param}
\resizebox{0.47\textwidth}{!}{%
\begin{tabular}{cl|ccl|c}
\cline{1-3} \cline{5-6}
Modality & \multicolumn{1}{c|}{Encoder Type} & \#Param &  & \multicolumn{1}{c|}{Module} & \#Param \\ \cline{1-3} \cline{5-6} 
m1 & PointPillar & 0.23M &  & Adapter & 0.30M \\
m2 & Second & 0.97M &  & Codebook & 0.02M \\
m3 & VoxelNet & 0.56M &  & Translator & 0.12M \\
m4 & PointPillar & 0.23M &  & Backend & 5.27M \\
m5 & PointPillar & 0.23M &  & - Pyramid Fusion & 3.79M \\
m6 & LSS(ResNet-101) & 1.77M &  & - Shrink Net & 1.48M \\
m7 & LSS(EfficientNet) & 14.95M &  & - Heads & 5k \\ \cline{1-3} \cline{5-6} 
\end{tabular}%
}
\end{table}

Our setting accommodates a wide spectrum of heterogeneous modalities, including different sensor types (which is the most influential), diverse encoder networks, and LiDARs with different beam counts and varying voxel size. These differences lead to substantially divergent intermediate feature representations, making direct fusion challenging. Notably, encoder sizes vary dramatically, especially for vision-based modalities like m7, rendering full retraining of all encoders impractical in terms of computation requirements. In contrast, our alignment components (Adapter: 0.30M, Codebook: 0.02M, Translator: 0.12M) are extremely lightweight, easy to expand on a large scale. Moreover, the backend is frozen during code space construction, which drastically reduces training overhead while preserving strong detection capability. This modular and parameter-efficient strategy enables scalable, plug-and-play collaboration across highly heterogeneous sensing setups.

\section{Detailed Experiments on CodeAlign}
\label{sec: more exp}
\subsection{Code Space Construction} 

\begin{table}[h]
\centering
\setlength{\tabcolsep}{3pt}
\caption{Performance of different strategy in forming code space. Training methods include e2e training(E2E) and fix encoder and backbone training(Fix E\&B). TP is training parameter. }
\label{tab:code space construction}
\resizebox{0.47\textwidth}{!}{%
\begin{tabular}{lc|cccc|cccc}
\hline
\multicolumn{1}{c}{\multirow{2}{*}{Training}} & \multirow{2}{*}{Adapter} & \multicolumn{4}{c|}{m1} & \multicolumn{4}{c}{m6} \\
\multicolumn{1}{c}{} &  & AP30 & AP50 & AP70 & TP/M & AP30 & AP50 & AP70 & TP/M \\ \hline
E2E & \xmark & 95.07 & 94.64 & \textbf{91.05} & 5.50 & 55.08 & 45.51 & 25.19 & 7.04 \\
Fix E\&B & \xmark & 93.69 & 93.05 & 85.54 & 0.02 & 58.71 & 49.89 & 29.51 & 0.02 \\
Fix E\&B & \cmark & \textbf{95.2} & \textbf{94.65} & 89.37 & 0.30 & \textbf{58.85} & \textbf{50.30} & \textbf{31.10} & 0.30 \\ \hline
\end{tabular}%
}
\end{table}

Table~\ref{tab:code space construction} shows the effect of different code space setups. Freezing the encoder and backend (instead of end-to-end training) reduces trainable parameters and even improves performance. For LiDAR (\(m_1\)), CodeAlign boosts AP30 and AP50 but slightly drops AP70 by 1.68\%, likely because the fixed codebook doesn’t align perfectly with the fusion module, hurting spatial accuracy. For camera (\(m_6\)), it largely improves AP70 by 5.91\%, as the codebook captures cleaner and more compact visual features.

\begin{table}[h]
\centering
\caption{More ablation studies on code space construction of a single modality.}
\label{tab:detail code space}
\resizebox{0.47\textwidth}{!}{%
\begin{tabular}{cc|ccc|ccc}
\hline
\multicolumn{2}{c|}{Method} & \multicolumn{3}{c|}{m1} & \multicolumn{3}{c}{m6} \\
Strategy & Adapter & AP30 & AP50 & AP70 & AP30 & AP50 & AP70 \\ \hline
E2E & \xmark & 95.07 & 94.64 & 91.05 & 55.08 & 45.51 & 25.19 \\ \hline
\multirow{4}{*}{Fix Encoder} & \xmark & 95.36 & 94.57 & 85.34 & 59.39 & 50.85 & 31.34 \\
 & 4×ResBlock & 94.78 & 94.52 & \textbf{91.17} & 59.13 & 51.09 & \textbf{33.04} \\
 & 1×ConvNeXt block & \textbf{95.62} & \textbf{95.14} & 90.4 & 59.37 & 51.25 & 31.48 \\
 & 2-layer MLP & 95.45 & 94.91 & 89.67 & \textbf{59.4} & \textbf{51.29} & 32.23 \\ \hline
\multirow{4}{*}{\begin{tabular}[c]{@{}c@{}}Fix Encoder\\ \&Backend\end{tabular}} & \xmark & 93.69 & 93.05 & 85.54 & 58.71 & 49.89 & 29.51 \\
 & 4×ResBlock & \textbf{95.2} & \textbf{94.65} & \textbf{89.37} & 58.85 & 50.3 & \textbf{31.1} \\
 & 1×ConvNeXt block & 95.13 & 94.63 & 88.59 & \textbf{59.4} & \textbf{50.46} & 30.94 \\
 & 2-layer MLP & 94.40 & 93.65 & 86.03 & 59.11 & 50.2 & 29.71 \\ \hline
\end{tabular}%
}
\end{table}

\begin{table*}[t]
\centering
\caption{Ablation on input source and output type of translator.}
\label{tab:ablation translator}
\resizebox{0.7\textwidth}{!}{%
\begin{tabular}{lc|ccc|ccc|c}
\hline
\multicolumn{1}{c}{\multirow{2}{*}{Source}} & \multirow{2}{*}{Type} & \multicolumn{3}{c|}{m1 + m2} & \multicolumn{3}{c|}{m1 + m6} & \multirow{2}{*}{Comm Load} \\
\multicolumn{1}{c}{} &  & AP30 & AP50 & AP70 & AP30 & AP50 & AP70 &  \\ \hline
Encoded Feature & D2D & \textbf{95.65} & \textbf{95.06} & \textbf{90.72} & \textbf{88.93} & \textbf{86.96} & \textbf{75.04} & 32MB \\
Encoded Feature & D2C & 95.44 & 94.88 & 89.83 & 88.13 & 86.41 & 73.47 & 0.03MB \\
Adapted Feature & D2C & 95.49 & 94.9 & 90.11 & 88.28 & 86.42 & 73.56 & 0.03MB \\
Reconstructed Feature & D2C & 93.96 & 93.23 & 85.87 & 76.13 & 73.42 & 53.89 & 0.03MB \\
Code Map & C2C & 94.58 & 93.87 & 87.4 & 76.93 & 74.26 & 55.16 & 0.03MB \\ \hline
\end{tabular}%
}
\end{table*}

Table~\ref{tab:detail code space} presents a more detailed ablation study, including variants that only fix the encoder and comparisons of different adapter designs. The experiments show that fixing only the encoder while allowing the backend to be trained yields slightly better performance, as the backend can adapt to the relatively discrete features produced by the codebook; notably, AP70 even surpasses that of full end-to-end training in some cases, likely because the discretized representations suppress certain noise. In contrast, fixing the backend incurs a modest drop of nearly 2\% in AP70 but dramatically reduces training cost, making the approach more scalable. Regarding adapter architectures—ResBlock, ConvNeXt, and MLP—their parameter counts decrease progressively. As shown in the table, lightweight adapters already achieve satisfactory AP30 and AP50 scores, but for fine-grained localization (AP70), a larger ResBlock-based adapter remains necessary to properly align cross-modal representations.

\begin{table}[h]
\centering
\caption{Single-modality performance comparison between end-to-end Pyramid Fusion and code space construction.}
\label{tab:single code space}
\resizebox{0.47\textwidth}{!}{%
\begin{tabular}{c|ccc|ccc}
\hline
\multirow{2}{*}{Modality} & \multicolumn{3}{c|}{E2E Pyramid Fusion} & \multicolumn{3}{c}{Code Space Construction} \\
   & AP30  & AP50  & AP70  & AP30  & AP50  & AP70  \\ \hline
m1 & 94.69 & 94.14 & 90.29 & 95.2  & 94.65 & 89.37 \\
m2 & 94.93 & 94.54 & 91.72 & 94.74 & 94.31 & 90.23 \\
m3 & 95.86 & 95.46 & 91.51 & 95.00 & 94.47 & 88.89 \\
m4 & 89.26 & 88.66 & 83.68 & 86.79 & 85.82 & 77.47 \\
m5 & 95.02 & 94.60 & 89.94 & 94.71 & 93.57 & 87.96 \\
m6 & 58.11 & 50.79 & 33.72 & 58.85 & 50.30 & 31.10 \\
m7 & 61.93 & 54.27 & 37.17 & 61.93 & 54.22 & 35.57 \\ \hline
\end{tabular}%
}
\end{table}

To comprehensively evaluate the code space construction across all modalities, we conduct the experiments shown in Table~\ref{tab:single code space}. As demonstrated, for every modality, our proposed plug-in manner code space construction method achieves comparable perception performance to the end-to-end trained Pyramid Fusion, while explicitly extracting a codebook-based feature representation. This approach introduces only minimal additional training overhead and reduces communication cost by orders of magnitude.

\subsection{Feature-Code-Feature Translation}

\begin{table}[h]
\centering
\caption{Performance of more modalities aligned to m1 when m1 is ego with feature-code-feature translation.}
\label{tab:modality translation}
\resizebox{0.3\textwidth}{!}{%
\begin{tabular}{c|ccc}
\hline
Scenario & AP30 & AP50 & AP70 \\ \hline
m1 single & \multicolumn{1}{l}{81.18} & \multicolumn{1}{l}{79.44} & \multicolumn{1}{l}{68.26} \\
m1 + m2 & 95.49 & 94.90 & 90.11 \\
m1 + m3 & 95.22 & 94.67 & 89.48 \\
m1 + m4 & 94.63 & 93.93 & 87.61 \\
m1 + m5 & 95.06 & 94.53 & 89.23 \\
m1 + m6 & 88.13 & 86.41 & 73.47 \\
m1 + m7 & 88.11 & 86.59 & 74.47 \\ \hline
\end{tabular}%
}
\end{table}

Table~\ref{tab:modality translation} shows the performance of various modalities aligned to m1 via feature-code-feature translation. Despite significant heterogeneity in sensor types and encoder architectures, all agents consistently enhance perception performance over the m1-single baseline. Notably, collaborations with LiDAR-based neighbors (m2–m5) yield substantial gains (e.g., +14.31 AP30 with m2), while even camera-based modalities (m6, m7) provide meaningful improvements (+7.0 AP30). These results demonstrate that feature-code-feature translation enables diverse agents to participate effectively in cooperative perception with minimal communication overhead, significantly boosting accuracy without requiring retraining of existing components.

\begin{table}[h]
\centering
\caption{Feature-code-feature translation can be trained under single-agent data. We show an example of scene m1+m2.}
\label{tab:single training}
\resizebox{0.34\textwidth}{!}{%
\begin{tabular}{l|ccc}
\hline
Data Type          & AP30  & AP50  & AP70  \\ \hline
Collaborative Data & 95.44 & 94.88 & 89.83 \\
Single Data        & 94.83 & 94.24 & 88.09 \\ \hline
\end{tabular}%
}
\end{table}

A key advantage of our framework is that the code translator can be trained using only single-agent (non-collaborative) data, eliminating the need for synchronized multi-vehicle scenes during training. This greatly enhances practicality in real-world settings where collaborative data is scarce or unavailable. Table~\ref{tab:single training} validates this capability. When trained solely on single-agent data, the translator achieves 94.83 AP30 and 94.24 AP50. Compared to training with collaborative data, it leads to a performance drop of less than 1 in AP50. This minor degradation demonstrates that high-quality cross-modal alignment can be learned from local perception data alone, making our method highly adaptable to diverse deployment scenarios.

\begin{table*}[t]
\centering
\caption{Ablation on different codebook sizes.}
\label{tab:codebook size}
\resizebox{0.7\textwidth}{!}{%
\begin{tabular}{c|ccc|ccc|ccc}
\hline
\multirow{2}{*}{Codebook Size} & \multicolumn{3}{c|}{m1} & \multicolumn{3}{c|}{m6} & \multicolumn{3}{c}{m1 + m6} \\
 & AP30 & AP50 & AP70 & AP30 & AP50 & AP70 & AP30 & AP50 & AP70 \\ \hline
4 & 94.01 & 93.37 & 84.80 & 56.88 & 47.90 & 26.84 & 84.49 & 82.48 & 66.30 \\
8 & 94.85 & 94.24 & 87.94 & 58.07 & 49.36 & 30.07 & 86.86 & 84.59 & 69.49 \\
16 & 95.20 & 94.65 & 89.37 & 58.85 & 50.30 & 31.10 & 88.13 & 86.41 & 73.47 \\
32 & 95.02 & 94.42 & 89.09 & 58.56 & 49.84 & 29.20 & 88.23 & 86.08 & 73.82 \\
64 & 95.18 & 94.72 & 89.75 & 59.02 & 50.55 & 31.93 & 87.75 & 85.98 & 74.56 \\ \hline
\end{tabular}%
}
\end{table*}

Table~\ref{tab:ablation translator} analyzes the impact of input source and output type on translation performance. The D2D (dense-to-dense) variant achieves the highest accuracy by preserving full feature fidelity, but incurs a prohibitive communication cost of 32MB, failing to address the core bottleneck. In contrast, D2C (dense-to-code) and C2C (code-to-code) translation drastically reduce communication to just 0.03MB by transmitting compact code maps. However, C2C suffers severe performance degradation (AP70 drop by nearly 20 for m1+m6) due to excessive information loss in direct code-to-code mapping. Among D2C variants, using raw encoded features yields strong results, while reconstructed features suffer significant drops like C2C with similar reason. Notably, adapted features achieve better performance, slightly outperforming raw encoded features and confirming that adapter enhances cross-modal alignment.

Table~\ref{tab:codebook size} shows the effect of codebook size on perception performance. As the codebook size grows from 4 to 16, AP metrics consistently improve for all settings, indicating that a larger codebook captures richer and more discriminative semantic information. Beyond size 16, however, performance plateaus, suggesting that 16 entries are sufficient to represent the essential environmental semantics in this setting. Moreover, the mapping difficulty increases with codebook size growing, where bigger codebook even leads to lower translation performance. Thus, a codebook size of 16 achieves the optimal trade-off between representation capacity and communication efficiency.

\subsection{Group Code Space Construction}

\begin{table}[h]
\centering
\caption{More ablation studies on group code space construction, especially on similarity losses, in group (m1,m6).}
\label{tab:sim ablation}
\resizebox{0.47\textwidth}{!}{%
\begin{tabular}{l|lccc}
\hline
\multicolumn{1}{c|}{Strategy}                                                        & \multicolumn{1}{c}{$L_{sim}$} & AP30  & AP50  & AP70  \\ \hline
E2E & \textbackslash{} & 89.3           & \textbf{88.03} & \textbf{80.44} \\
+ Codebook(16) & \textbackslash{} & \textbf{89.54} & 87.74          & 77.87          \\ \hline
\multirow{9}{*}{\begin{tabular}[c]{@{}l@{}}++ Fix Enc\\ ++ Add Adapter\end{tabular}} & \textbackslash{}              & 88.37 & 86.83 & 78.09 \\
               & L2               & 88.82          & 87.31          & 78.88          \\
               & Instance         & 88.99          & 87.01          & 76.08          \\
               & Reconstruction   & 88.89          & 87.36          & 79.06          \\
               & Cosine           & 88.66          & 87.29          & 79.54          \\
               & Smooth L1        & \textbf{89.04} & \textbf{87.54} & 79.63          \\
               & MMD              & 88.76          & 87.28          & \textbf{79.75} \\
               & Coral            & 88.73               & 87.23               & 79.03               \\
               & JS               & 89.01               & 87.50               & 79.14               \\ \hline
\end{tabular}%
}
\end{table}

Table~\ref{tab:sim ablation} presents ablation studies on group code space construction. The end-to-end Pyramid Fusion baseline achieves the highest performance, as it incurs no information loss. Introducing a codebook alone preserves coarse-level accuracy but degrades fine-grained localization (AP50 drops by 0.29, AP70 by 2.57), indicating quantization harms spatial detail. By freezing the encoder, adding an adapter, and incorporating a similarity loss $L_{\text{sim}}$, performance across all metrics improves significantly, nearly recovering the baseline while enabling compact feature representation. We evaluate multiple similarity losses; all consistently enhance alignment over the no-loss variant. Smooth L1 Loss yields the best overall results, facilitating stable and precise alignment of heterogeneous feature distributions.

\subsection{Group Feature-Code-Feature Translation}

\begin{table}[h]
\centering
\caption{Alignment performance between group (m1,m6) and group (m2,m7) in order [m1,m7,m2,m6].}
\label{tab:group_inter_1}
\resizebox{0.4\textwidth}{!}{%
\begin{tabular}{l|cccc}
\hline
 & AP30 & AP50 & AP70 & Comm Load \\ \hline
Late Fusion & 89.61 & 86.47 & 69.97 & 0.5KB \\
HEAL & 92.50 & 91.74 & 85.64 & 32MB \\
CodeAlign & 92.27 & 91.48 & 85.42 & 0.03MB \\ \hline
\end{tabular}%
}
\end{table}

Table~\ref{tab:group_inter_1} demonstrates CodeAlign’s effectiveness in cross-group alignment. Here, modalities m1 and m6 share a common codebook, and the translator aligns m2, m7 into this common code space. Despite the heterogeneity across groups, CodeAlign achieves performance on par with HEAL, while reducing communication overhead. This validates that inter-group alignment via a shared discrete code space is both feasible and efficient.

\begin{table}[h]
\centering
\caption{HEAL degrades the alignment between the original group after being aligned to other groups. In order [m2,m7]}
\label{tab:group_inter_2}
\resizebox{0.4\textwidth}{!}{%
\begin{tabular}{l|lll}
\hline
                      & AP30  & AP50  & AP70  \\ \hline
Late Fusion           & 75.25 & 69.00 & 50.20 \\
HEAL(pyramid fusion)  & \textbf{85.73} & \textbf{84.21} & \textbf{76.41} \\
HEAL(backward alignment) & 83.76 & 82.55 & 73.59 \\ \hline
\end{tabular}%
}
\end{table}

\begin{table*}[t]
\centering
\caption{Comparison of heterogeneous collaborative perception methods in terms of encoder/sensor diversity, extensibility, and adaptability to Modality Isolation. Remarks detail limitations regarding isolation handling and key drawbacks.}
\label{tab:compare method}
\setlength{\tabcolsep}{3pt}
\resizebox{\textwidth}{!}{%
\begin{tabular}{l|cccc|l}
\hline
Method & Dif Enc & Dif Sensor & Extensible & M.I. Adapt & Remarks \\ \hline
V2X-ViT~\cite{v2xvit} & \cmark &  &  &  & E2E attention fusion; needs same-scene features for spatial attention. \\
HM-ViT~\cite{HM-ViT} &  & \cmark &  &  & E2E attention fusion; needs same-scene features for spatial attention. \\
MPDA~\cite{xu2022model} & \cmark &  &  &  & E2E attention fusion; needs same-scene features for spatial attention. \\
PnPDA~\cite{luo2024plug} & \cmark &  &  &  & Contrastive loss for hetero features requires co-occurring modalities. \\
PolyInter~\cite{xia2025one} & \cmark &  & \cmark &  & Channel selection and spatial attention module need co-occurring modalities. \\
STAMP~\cite{gao2025stamp} & \cmark & \cmark & \cmark &  & Aligns to protocol net; protocol modality must be present in every scene. \\
CodeFilling~\cite{hu2024communication} & \cmark & \cmark &  &  & All modalities needed to train shared codebook; degrades with isolated modalities. \\
HEAL~\cite{Heal} & \cmark & \cmark & \cmark & \cmark & Supports modality isolation; but needs encoder retraining with high training cost. \\
CodeAlign & \cmark & \cmark & \cmark & \cmark & Supports modality isolation; plug-in design; training-efficient, communication-efficient. \\ \hline
\end{tabular}%
}
\end{table*}

Table~\ref{tab:group_inter_2} highlights a critical limitation of HEAL. Specifically, HEAL requires retraining encoders of non-base groups to align them with a single base group, disrupting the original intra-group alignment. This process degrades the internal consistency and performance within those groups. For instance, when m2 and m7 are realigned to the base group (m1, m6), their mutual alignment drop in AP70 by 2.82, compared to the original pyramid fusion performance.

\section{Comparison to Other Methods}
\label{sec: compare method}
The comparison in Table~\ref{tab:compare method} highlights fundamental limitations of existing heterogeneous collaborative perception methods in handling modality isolation and practical deployment. V2X-ViT, HM-ViT, and MPDA all rely on end-to-end attention-based fusion, which fundamentally assumes that input features from different agents are spatially aligned, which is a condition only satisfied when modalities co-occur in the same scene during training. Similarly, PnPDA employs contrastive loss to align heterogeneous features, but this alignment also requires paired, spatially corresponding data from multiple modalities in identical scenarios. PolyInter leverages channel selection and spatial attention modules, yet these components still depend on co-occurring modalities to learn meaningful cross-agent interactions.

STAMP introduces a protocol network to unify representations, but it mandates that the protocol modality be present in every scene so that other modalities can align through feature similarity, making it inapplicable when the protocol agent is absent. CodeFilling attempts to learn a shared codebook across all modalities, but this requires simultaneous access to all modalities during training; consequently, its performance degrades significantly under modality isolation, and it lacks extensibility to new sensors without retraining the entire system.

HEAL addresses modality isolation via backward alignment, retraining each encoder locally to map into a common space. While effective, this approach incurs high computational cost and operational inconvenience where every new modality or update necessitates full encoder retraining. In contrast, our method, CodeAlign, decouples representation learning by constructing modality-specific code spaces and establishing a lightweight feature-code-feature translation pipeline. This design eliminates the need for co-occurring data during training, supports plug-and-play integration of new modalities, and enables both training-efficient adaptation and ultra-low communication overhead. These advantages make it uniquely suited for real-world heterogeneous multi-agent systems under dynamic and incomplete observation conditions.

\end{document}